\documentclass[conference]{IEEEtran}
\IEEEoverridecommandlockouts

\usepackage{cite}
\usepackage{microtype}
\usepackage{graphicx}
\usepackage{subfigure}
\usepackage{booktabs} 
\usepackage{hyperref}
\usepackage{amsmath,amssymb,amsfonts}
\usepackage{mathtools}
\usepackage{amsthm}
\usepackage{algorithm}
\usepackage{algorithmic}
\usepackage{xcolor}
\usepackage{natbib}
\theoremstyle{plain}

\theoremstyle{definition}

\theoremstyle{remark}

\title{Performant LLM Agentic Framework for Conversational AI}

\title{Performant LLM Agentic Framework for Conversational AI}

\author{
\IEEEauthorblockN{Alex Casella\IEEEauthorrefmark{1}}
\IEEEauthorblockA{Thoughtly, Boston University\\
Email: alex@alexcasella.io}
\and
\IEEEauthorblockN{Wayne Wang\IEEEauthorrefmark{1}}
\IEEEauthorblockA{Thoughtly, UC Berkeley\\
Email: haohanw@eecs.berkeley.edu}
\thanks{\IEEEauthorrefmark{1} Equal Contribution.}
}

\begin{document}

\maketitle

\begin{abstract}
The rise of Agentic applications and automation in the Voice AI industry has led to an increased reliance on Large Language Models (LLMs) to navigate graph-based logic workflows composed of nodes and edges. However, existing methods face challenges such as alignment errors in complex workflows and hallucinations caused by excessive context size. To address these limitations, we introduce the Performant Agentic Framework (PAF), a novel system that assists LLMs in selecting appropriate nodes and executing actions in order when traversing complex graphs. PAF combines LLM-based reasoning with a mathematically grounded vector scoring mechanism, achieving both higher accuracy and reduced latency. Our approach dynamically balances strict adherence to predefined paths with flexible node jumps to handle various user inputs efficiently. Experiments demonstrate that PAF significantly outperforms baseline methods, paving the way for scalable, real-time Conversational AI systems in complex business environments.
\end{abstract}


\section{Introduction}
\label{Introduction}
Graph-based workflows are central to numerous business processes across industries such as education, legal, healthcare, and customer support. These workflows represent decision-making steps as nodes, and connections between them as edges. The rise of Conversational AI within these spaces introduces new challenges. Autonomous agents, powered by large language models (LLMs), are increasingly being used to navigate these workflows, enabling the automation of complex business processes \citep{zhuge2023aflow}. Each node in the workflow contains specific instructions or prompts that guide the agent’s speech generation and certain actions to trigger. Nodes can be classified into several types, including Start Nodes, which define the root and entry point of a workflow; End Nodes, which signal the termination of the workflow; and generic Nodes, which serve as intermediate decision points containing speech instructions for the LLM to converse with users in predefined ways. Additionally, Transfer Nodes in Conversational AI workflows allow for transitioning the conversation to another autonomous or human agent. Edges between nodes may include logical conditions that dictate the agent’s transitions, ensuring workflows are executed accurately.

\begin{figure}[ht]
\centering
\includegraphics[width=\columnwidth]{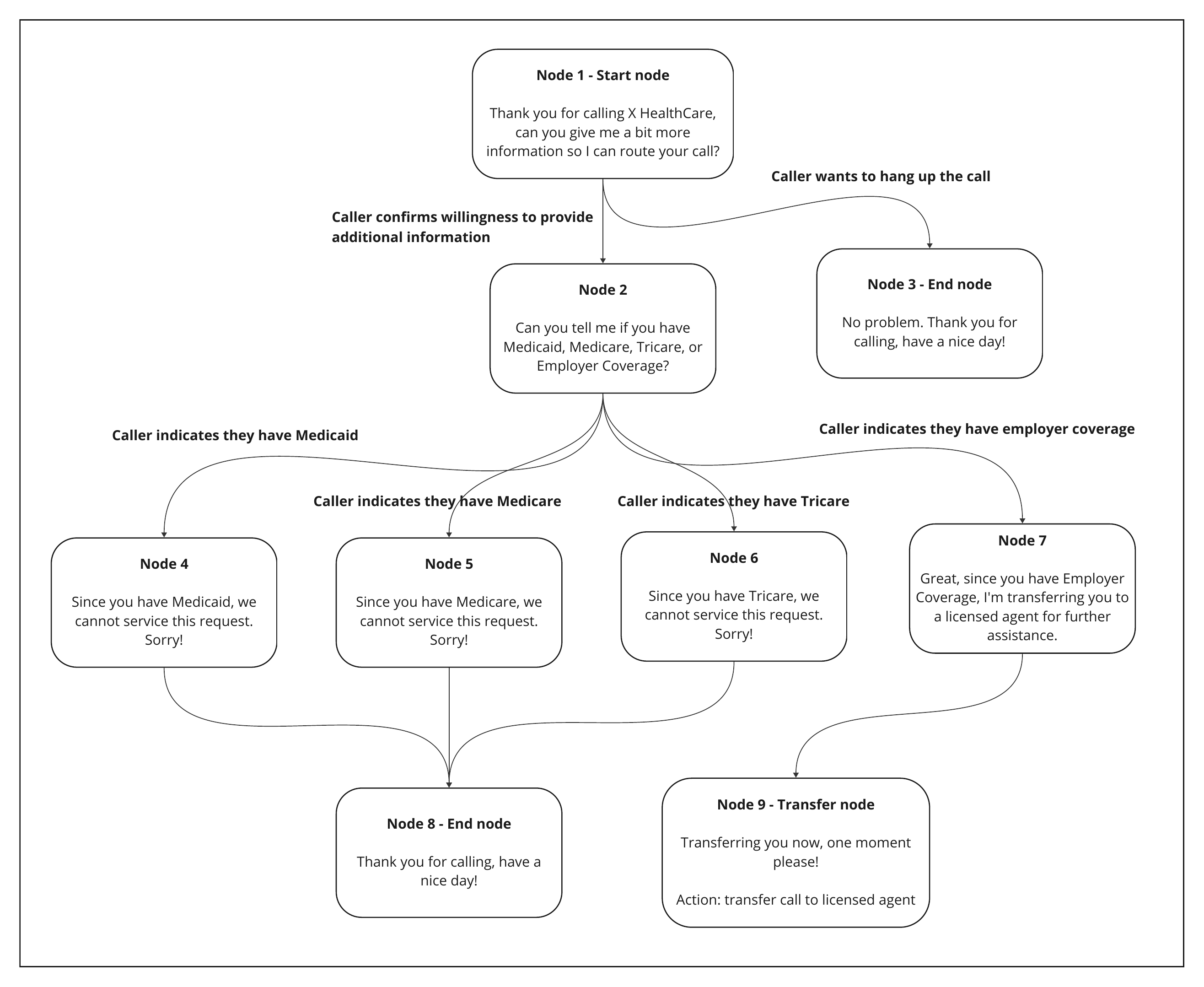}
\caption{Example illustration of an Agentic workflow for a healthcare call center use case, where the Agent needs to route calls based on different conditions.}
\label{exmaple-call-1}
\end{figure}

\begin{figure}[ht]
\centering
\includegraphics[width=\columnwidth]{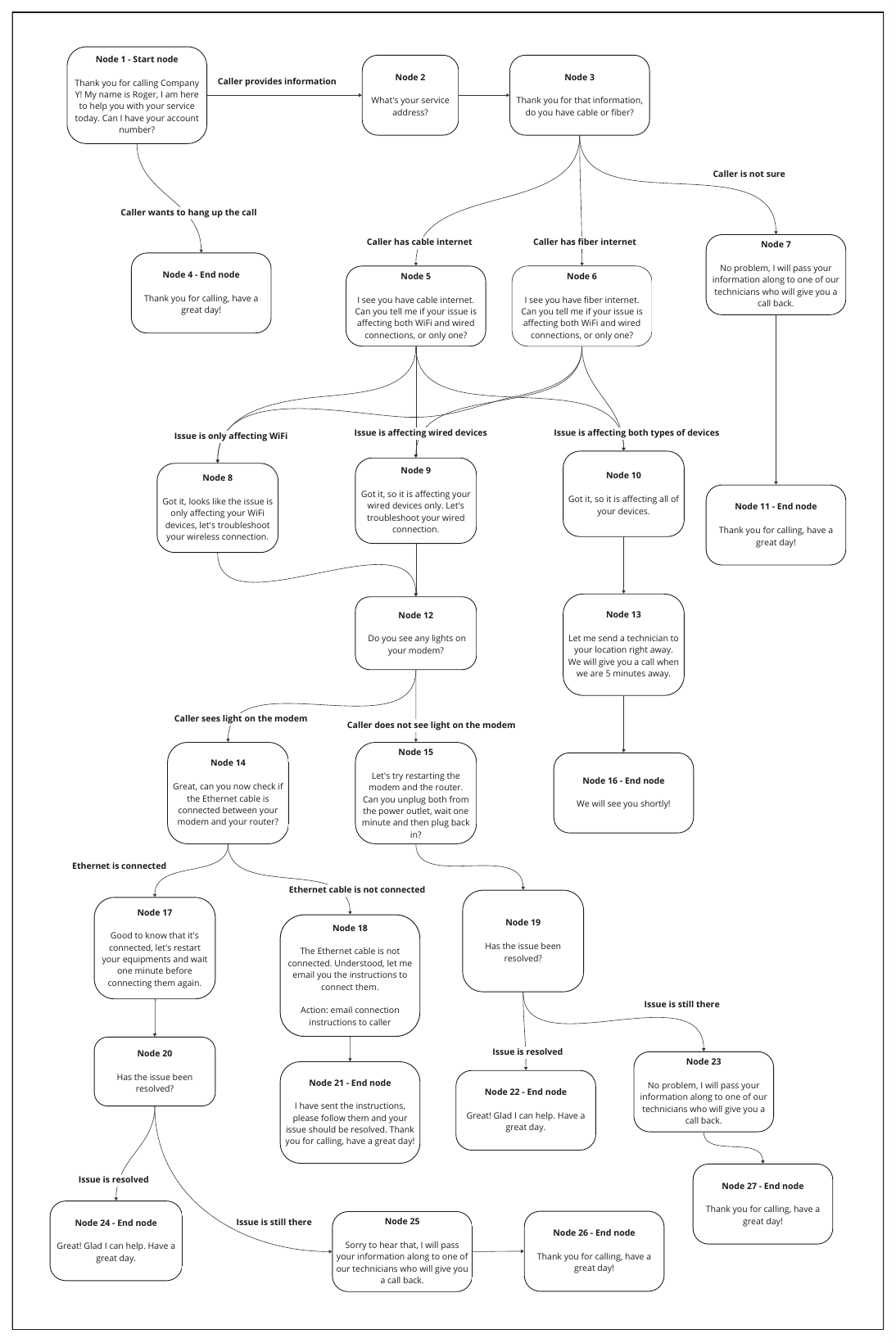}
\caption{Example illustration of an Agentic workflow for an internet service company helping callers troubleshoot connection issues. This workflow demonstrates how a more complex use case can have more conditions, nodes, and edges.}
\label{exmaple-call-2}
\end{figure}

Figure~\ref{exmaple-call-1} illustrates how tasks such as determining health care eligibility can be broken down into nodes, edges, and conditions. For example, a healthcare provider might use such a workflow to efficiently filter out patients without the required insurance, reducing the burden on human agents. However, workflows like these can rapidly grow in complexity. As shown in Figure~\ref{exmaple-call-2}, adding just a few additional conditions to the conversation flow can drastically increase the number of nodes and edges, making the workflow more difficult to manage and execute effectively.

Although LLMs such as GPT and LLAMA are built on autoregressive decoder-based transformer architectures optimized for natural language generation, they are not inherently designed to handle structured, multi-step processes with extensive context \citep{qiu2024chatgpt, shi2023bert}. Existing approaches have been to add a planning phase, where the LLM would take time to orchestrate the action, and then proceed to the generation tasks \citep{valmeekam2023planning, zhou2024isrllm}. However, this approach is not optimal to the Conversational AI use case, as it would increase the overall latency by doubling the number of queries needed. Tasks such as managing end-to-end customer service requests with non-standard return policies, performing outbound sales calls that involve dynamic CRM updates, or redirecting users to appropriate departments after a sequence of filtering questions require precision, alignment, and low-latency responses. These limitations force businesses to oversimplify workflows, sacrificing accuracy and operational efficiency—an outcome that is contrary to their objectives.

The challenges inherent in adapting LLMs to graph-based workflows underscore the need for new approaches that can accurately and efficiently execute workflows while respecting real-world constraints such as latency. While adding more reasoning steps could theoretically improve accuracy, such methods are impractical for Conversational AI applications where rapid response times are critical.

To address these challenges in the current Conversational AI space, this paper introduces the \textbf{Performant Agentic Framework (PAF)}, a novel solution for efficient graph traversal that balances accuracy and latency in real-world applications. By leveraging both traditional decision-making logic and mathematical methods for next-node selection, PAF enables agents to execute workflows with greater precision and speed. Our experiments demonstrate that PAF significantly outperforms baseline and traditional methods in both accuracy and latency, as evidenced by higher alignment scores and reduced response times.

\section{Related Work}
\label{Related Work}
The reliance on LLM-based systems to execute graph-based workflows has seen significant research attention, particularly in developing frameworks that aim to balance accuracy, latency, and alignment with predefined workflows. Below, we discuss prominent related works and their limitations.

\textbf{Agentic Framework} Serving as examples, LangChain \citep{langchain2023} and LangGraph \citep{langgraph2023} streamline graph-based workflows by utilizing function calls and prompt chaining. While effective for simple tasks, their reliance on keyword-based triggers often results in alignment errors, especially in workflows with hundreds or thousands of nodes. These frameworks lack robustness for real-world applications where actions must be dynamically triggered at various points in conversations. Furthermore, their reliance on LLM-generated triggers leads to unreliability in critical workflows, where adherence to predefined paths is essential for compliance and business logic \citep{langchain2023, langgraph2023}. Additionally, limitations in LLM context windows further exacerbate their inefficiency in retaining relevant information across extended workflows, introducing hallucinations and context drift during execution \citep{dong2024hallucination}.

\textbf{Conversational AI} Conversational AI has been a key focus for Natural Language Processing. The existing Conversational AI solutions emphasize the need for multi-modality, guardrails, and advanced tuning to enhance dialogue quality \citep{dong2023nextgen}. Prior approaches to the Voice AI space have been proven to work in a sandbox conversational setting \citep{james2024voice}, but lack the consistency and accuracy required for production use. As suggested, LLMs miss certain abilities to maintain performance in a dynamic conversational setting, unable to handle numerous tasks conditionally while reducing hallucinations and staying within context \citep{gill2023chatgpt, dong2023nextgen, dong2024hallucination}. 

\textbf{MetaGPT and SOP Translation} MetaGPT leverages Standardized Operating Procedures (SOPs) to structure workflows, enabling agents to replicate domain-specific expertise. However, its reliance on iterative planning and validation increases latency, making it unsuitable for real-time applications. For example, as noted in \citep{gao2023metagpt}, the planning phase requires additional LLM calls, which adds computational overhead. While MetaGPT is effective for SOP alignment, it struggles with unusual user inputs and extended workflows, leading to significant context drift. Its dependence on domain-specific fine-tuning also hinders generalizability, limiting its use in broader applications \citep{gao2023metagpt, wang2024metagpt}.

\textbf{Comparison and Our Contributions} Existing frameworks have made valuable contributions but are hindered by issues such as context drift, high latency, and alignment errors. PAF addresses these limitations by replacing LLM planning phases with a mathematical decision-making approach, combining vector-based node selection and specialized prompt engineering. Unlike previous methods, PAF reduces context size while improving accuracy, making it a scalable and production-ready solution for navigating graph-based workflows.

\section{Performant Agentic Framework (PAF)}
\label{Performant Agentic Framework (PAF)}
PAF is a framework designed for Agentic workflows, enabling agents to navigate graph-based structures composed of nodes and edges to execute predefined workflows. It is comprised of two components: Basic PAF and Optimized PAF, each tailored to address specific challenges in workflow execution.

\subsection{Basic PAF}
\textbf{Problem Formulation} PAF enables agents to operate by following nodes connected by logical edges. During each generation turn, the agent follows the nodes in the graph according to the logical conditions specified as outcomes of the node. If a condition is met, the agent navigates to and executes the instructions of the next node in the graph. 

Our PAF involves leveraging LLM as a Judge for identifying the Agent's location in the map dynamically per each generation as follows:

\begin{algorithm}[ht]
\caption{LLM as a Judge for Node Identification}
\label{alg:judge_node_search}
\begin{algorithmic}
   \STATE {\bfseries Input:} $ConversationHistory$, $NavigationMap$, $LastestIdentifiedNode$
   \STATE {\bfseries Output:} $UpdatedLatestIdentifiedNode$
   \STATE {\bfseries Step 1: Format Input for the LLM}
   \STATE \hspace{1em} Construct a prompt using $ConversationHistory$
   \STATE \hspace{1em} Add a contextual anchor by traversing from the StartNode to $LastestIdentifiedNode$ and collect all first layer children Nodes in the map, e.g., "You were previously on Node \texttt{\{LastestIdentifiedNode\}} with options to navigate to in the map \texttt{\{Path\}} each with instructions being..."
   \STATE \hspace{1em} If $LastestIdentifiedNode$ is unavailable, use: "This is the start of the task \texttt{\{task\}}, proceed to Node 0."
   \STATE {\bfseries Step 2: Query the LLM}
   \STATE \hspace{1em} Send a question to the LLM: 
   \STATE \hspace{1em} "Based on your latest responses, where are you currently in the navigation map?"
   \STATE {\bfseries Step 3: Process the Response}
   \STATE \hspace{1em} Parse the response to identify the node mentioned by the LLM.
   \STATE \hspace{1em} Validate the identified node against $NavigationMap$.
   \STATE {\bfseries Step 4: Return the Result}
   \STATE \hspace{1em} Output the validated node as $UpdatedLatestIdentifiedNode$.
\end{algorithmic}
\end{algorithm}

This design is particularly effective in production AI systems as it separates the generation tasks from other downstream modules, like Text-to-Speech (TTS). This modular approach optimizes latency, enabling parallel processing by downstream services such as a TTS service. Compared to implementations where prompts are added in a single body, Basic PAF achieves lower error rates by using a step-by-step logic tree, reducing the need for additional validation iterations through customized testing schemas \citep{li2023workflow, reddy2021summarizer}.

\begin{algorithm}[ht]
\caption{Basic Agentic Framework}
\label{alg:agentic_framework}
\begin{algorithmic}
   \STATE {\bfseries Input:} $ConversationHistory$, $NavigationMap$, $LatestIdentifiedNode$
   \STATE {\bfseries Output:} $UpdatedLatestIdentifiedNode$
   \STATE {\bfseries Step 1: Initialize LLM Instructional Message}
   \STATE \hspace{1em} Construct an instructional prompt for the LLM agent.
   \STATE \hspace{1em} Add $ConversationHistory$ to the prompt in a formatted structure.
   \STATE \hspace{1em} Include instructions based on $LatestIdentifiedNode$, e.g., "You are currently on Node \texttt{\{LatestIdentifiedNode\}}."
   \STATE \hspace{1em} Constructed navigation prompt by traversing the $NavigationMap$ and collecting all first layer children nodes' instructions on the \texttt{\{Path\}} from StartNode to $LatestIdentifiedNode$.
   \STATE {\bfseries Step 2: Query the LLM}
   \STATE \hspace{1em} Send the query to the LLM: 
   \STATE \hspace{1em} "Based on the navigation map and your current node, respond to the user question: \texttt{\{user question\}}."
   \STATE {\bfseries Step 3: Process LLM Output in a Streaming Loop}
   \WHILE{LLM agent streams output}
      \STATE (a) Identify Current Node:
      \STATE \hspace{2em} Use \textbf{Algorithm \ref{alg:judge_node_search}} to determine the node the agent selects.
      \STATE (b) Update LatestIdentifiedNode:
      \STATE \hspace{2em} Set $LatestIdentifiedNode$ to the node identified in Step (a).
      \STATE (c) Trigger Actions:
      \STATE \hspace{2em} Execute any actions related to the newly identified $LatestIdentifiedNode$.
      \STATE (d) Update NavigationMap:
      \STATE \hspace{2em} Modify $NavigationMap$ as needed based on the new $LatestIdentifiedNode$.
   \ENDWHILE
   \STATE {\bfseries Step 4: Return the Updated Node}
   \STATE \hspace{1em} Output $UpdatedLatestIdentifiedNode = LatestIdentifiedNode$.
\end{algorithmic}
\end{algorithm}

\subsection{Optimized PAF}
While Basic PAF offers significant improvements (shown later in the experiment section), it faces bottlenecks when workflows involve numerous nodes (e.g., 50 nodes with 4 conditions each). These bottlenecks arise as the agent struggles to differentiate between semantically similar prompts on different paths of the graph. Optimized PAF addresses this with vector-based scoring to reduce the size of the context window and improve logical adherence.

The heart of optimized PAF is the \textbf{Vector Node Search}, which evaluates nodes using embedding models with a confidence threshold as follows:

\begin{algorithm}[ht]
\caption{Vector-Based Node Search}
\label{alg:vector_node_search}
\begin{algorithmic}
   \STATE {\bfseries Input:} $NavigationMap$, $LatestIdentifiedNode$, $Threshold$, $LatestAgentResponse$
   \STATE {\bfseries Output:} $UpdatedLatestIdentifiedNode$
   \STATE {\bfseries Step 1: Vectorize Instructions and Agent Response}
   \STATE \hspace{1em} Compute vector representations for:
   \STATE \hspace{2em} (a) $LatestIdentifiedNode$.
   \STATE \hspace{2em} (b) $LatestIdentifiedNode$'s children nodes.
   \STATE \hspace{2em} (c) $LatestAgentResponse$ from the LLM.
   \STATE {\bfseries Step 2: Compute Similarity Scores}
   \STATE \hspace{1em} Compare the vector representation of $LatestAgentResponse$ against:
   \STATE \hspace{2em} (a) Vector for $LatestIdentifiedNode$ instruction.
   \STATE \hspace{2em} (b) Vectors for instructions of child nodes.
   \STATE \hspace{1em} Compute similarity scores using a suitable metric (e.g., dot product similarity).
   \STATE {\bfseries Step 3: Identify the Best Matching Node}
   \STATE \hspace{1em} Select the node with the highest similarity score.
   \STATE \hspace{1em} If the score exceeds the $Threshold$, assign the corresponding node as $UpdatedLatestIdentifiedNode$.
   \STATE {\bfseries Step 4: Update and Return}
   \IF{A node is identified with a score above the $Threshold$}
      \STATE \hspace{1em} Update $LatestIdentifiedNode$ to $UpdatedLatestIdentifiedNode$.
      \STATE \hspace{1em} Return $UpdatedLatestIdentifiedNode$.
   \ELSE
      \STATE \hspace{1em} Return false (Use LLM as a Judge Approach).
   \ENDIF
\end{algorithmic}
\end{algorithm}

\begin{algorithm}[ht]
\caption{Optimized Agentic Framework}
\label{alg:optimized_agentic_framework}
\begin{algorithmic}
   \STATE {\bfseries Input:} $ConversationHistory$, $NavigationMap$, $LatestIdentifiedNode$, $Threshold$
   \STATE {\bfseries Output:} $UpdatedLatestIdentifiedNode$

   \STATE {\bfseries Step 1: Precompute Vectorized Instructions}
   \STATE \hspace{1em} Compute and store vectorized representations for instructions at all nodes in $NavigationMap$.

   \STATE {\bfseries Step 2: Format Input for the LLM}
   \STATE \hspace{1em} Construct a message including:
   \STATE \hspace{2em} (a) Formatted $ConversationHistory$.
   \STATE \hspace{2em} (b) Instructions for $LatestIdentifiedNode$.
   \STATE \hspace{2em} (c) A constructed navigation prompt by traversing $NavigationMap$ and collecting all first-layer children nodes' instructions on the \texttt{\{Path\}} from StartNode to $LatestIdentifiedNode$.

   \STATE {\bfseries Step 3: Query the LLM Agent}
   \STATE \hspace{1em} Send the constructed message to the LLM agent.

   \STATE {\bfseries Step 4: Process LLM Output in a Streaming Loop}
   \WHILE{LLM agent streams output}
      \STATE (a) Perform Vector-Based Node Search (Algorithm~\ref{alg:vector_node_search}).
      \IF{a node is successfully identified}
         \STATE \hspace{2em} Proceed to Step 5.
      \ELSE
         \STATE (b) Perform LLM as a Judge (Algorithm~\ref{alg:judge_node_search}).
      \ENDIF
   \ENDWHILE

   \STATE {\bfseries Step 5: Update Current Node}
   \STATE \hspace{1em} Set $LatestIdentifiedNode$ to the node identified in Step 4.

   \STATE {\bfseries Step 6: Trigger Actions and Update the Graph}
   \STATE \hspace{1em} (a) Execute any actions related to the updated $LatestIdentifiedNode$.
   \STATE \hspace{1em} (b) Modify $NavigationMap$ as needed.

   \STATE {\bfseries Step 7: Return the Updated Node}
   \STATE \hspace{1em} Output $UpdatedLatestIdentifiedNode = LatestIdentifiedNode$.
\end{algorithmic}
\end{algorithm}

Optimized PAF leverages vector-based reasoning, incorporating both semantic direction and magnitude through the dot product. Notably, when comparing different metrics to use as a similarity score, we found that the dot product is particularly effective for Conversational AI applications. This finding aligns with research by \citep{huang2021dotproduct}, which demonstrates the advantages of using dot product as a vector score over cosine similarity, where cosine similarity may produce ambiguous results by ignoring magnitude. This is particularly relevant when dealing with over-fitted domain jargon, where it is critical for the agent to differentiate between subtly distinct expressions that hold drastically different implications. This approach aligns well with emerging models like OpenAI's text-2-vec-3-small \citep{openai-text-2-vec}, which are tuned to reflect confidence alongside semantic direction.

\section{Experiment}
\label{Experiment}
To evaluate the effectiveness of PAF, we designed experiments to compare the performance of PAF with existing graph traversal and node selection methodologies. These experiments are designed to assess the latency, accuracy, and alignment of the framework across various workflows, particularly in Conversational AI applications.

\subsection{Experiment Setup}
The experiments utilize a simulated dataset generated to mimic real-world workflows. 

\textbf{Dataset Generation:} The experiment utilized a synthetic dataset generated to simulate real-world workflows. Each dataset entry contained:
\begin{itemize}
    \item \textbf{SystemPrompt:} A node navigation map with Agentic logic.
    \item \textbf{ConversationHistory:} Turn-by-turn interactions between the user and the agent.
    \item \textbf{GoldenResponse:} A pre-verified response audited through LLM-As-a-Judge and human evaluation, serving as the ground truth.
\end{itemize}

Conversations were executed by two agents (a "user" LLM and a "response" LLM), with a random turn length (6--10). Golden responses were derived from the corresponding node’s instruction and validated by humans.

\subsection{Framework Performance Evaluation}
We evaluated three frameworks under the following metrics:
\begin{itemize}
    \item \textbf{Semantic Similarity:} Alignment between the generated response and the golden response using OpenAI’s text-2-vec-3-small embedding model \citep{openai-text-2-vec}.
    \item \textbf{Total Complete Hit Rate:} Percentage of responses that achieved a similarity score above 0.97.
    \item \textbf{Mean and Median Similarity Scores:} Overall alignment performance.
\end{itemize}

\textbf{Frameworks Tested:}
\begin{enumerate}
    \item \textbf{Baseline:} A naive approach treating the entire conversation as a single prompt.
    \item \textbf{Basic PAF:} A step-by-step logic tree leveraging LLM-as-Judge (Algorithms~\ref{alg:judge_node_search} and \ref{alg:agentic_framework}).
    \item \textbf{Optimized PAF:} A vector-based approach (Algorithms~\ref{alg:judge_node_search}, \ref{alg:vector_node_search}, \ref{alg:optimized_agentic_framework}).
\end{enumerate}

\subsection{Hypotheses}
\begin{itemize}
    \item \textbf{H1:} Basic PAF achieves higher average similarity than Baseline.
    \item \textbf{H2:} Optimized PAF achieves higher average similarity than Baseline.
    \item \textbf{H3:} Optimized PAF achieves higher average similarity than Basic PAF.
\end{itemize}
A one-sided paired t-test with $\alpha = 0.05$ was used for statistical significance.

\subsection{Experiment Steps}
\begin{enumerate}
    \item Simulate 100 conversations using the predefined workflow.
    \item Generate responses for each method.
    \item Compute similarity scores against the golden responses.
    \item Aggregate metrics such as total hit rate, mean, and median.
    \item Perform hypothesis testing.
\end{enumerate}

\subsection{Results}
\begin{table}[ht]
\caption{Result Metrics Across Algorithms}
\label{tab:metrics}
\centering
\begin{tabular}{lcccc}
\toprule
\textbf{Method} & \textbf{Total Hits} & \textbf{Count above 0.8} & \textbf{Mean} & \textbf{Median}\\
\midrule
Baseline & 0  & 0  & 0.391 & 0.387 \\
Basic PAF & 16 & 14 & 0.481 & 0.400 \\
Optimized PAF & 35 & 22 & 0.594 & 0.496 \\
\bottomrule
\end{tabular}
\end{table}

\begin{figure}[ht]
\centering
\includegraphics[width=0.8\columnwidth]{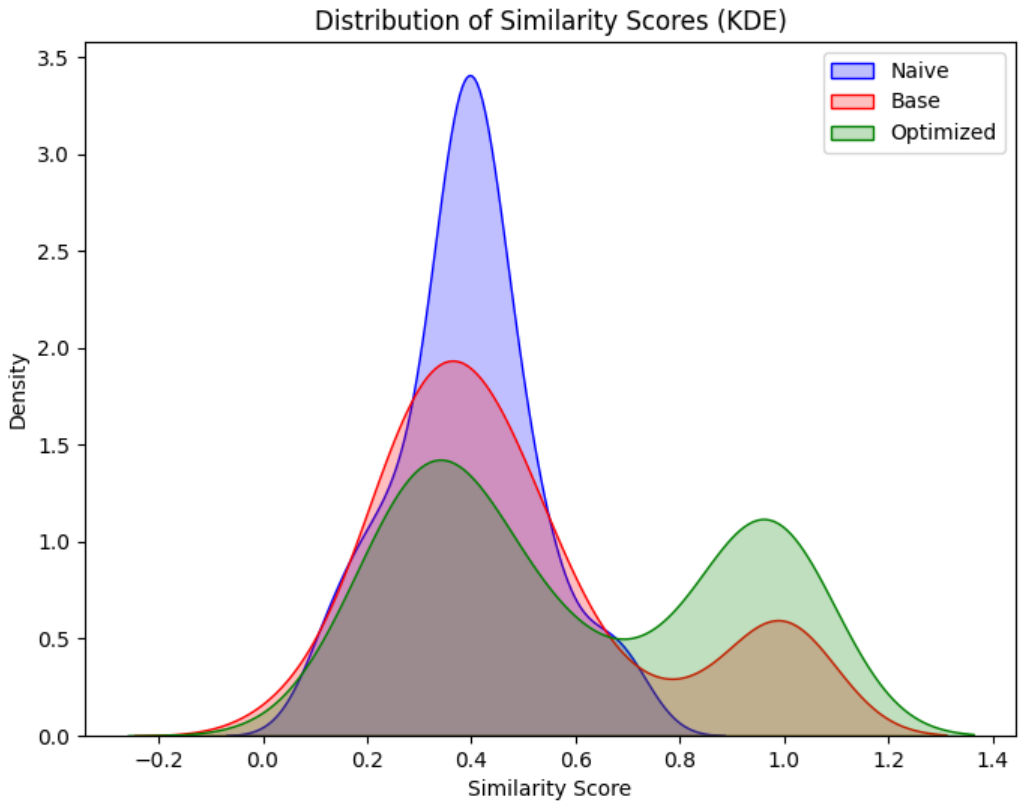}
\caption{Distribution of Similarity Scores for the 3 frameworks (Naive=Baseline, Base=Basic PAF, Optimized=Optimized PAF).}
\end{figure}

\begin{table}[ht]
\caption{Statistical Comparison Results (One-Sided Paired t-Tests)}
\label{statistical-results}
\centering
\begin{tabular}{lcc}
\toprule
\textbf{Comparison} & \textbf{t-statistic} & \textbf{p-value}\\
\midrule
Baseline vs Basic PAF & 2.9982 & 0.0020 \\
Baseline vs Optimized PAF & 7.3077 & 0.0000 \\
Basic PAF vs Optimized PAF & 4.2494 & 0.0000 \\
\bottomrule
\end{tabular}
\end{table}

\begin{itemize}
    \item \textbf{H1:} Basic PAF significantly outperforms the Baseline ($p=0.002$).
    \item \textbf{H2:} Optimized PAF significantly outperforms the Baseline ($p<0.001$).
    \item \textbf{H3:} Optimized PAF significantly outperforms Basic PAF ($p<0.001$).
\end{itemize}

\subsection{Reproducibility}
We provide code and data generation scripts in an anonymized repository.\footnote{\url{https://anonymous.4open.science/r/performant-agentic-framework-F5F6/README.md}}

\section{Conclusion}
Our approach introduces novel mechanisms for leveraging LLMs to navigate graph-based workflows, replacing the need for extensive planning phases and minimizing error rates. PAF achieves faster response times and greater accuracy in real-world applications by reducing reliance on large context windows and optimizing computational steps.

In summary, PAF resolves key limitations in existing Agentic frameworks by:
\begin{itemize}
    \item Removing extra iterations for validation and planning, thereby reducing latency.
    \item Improving alignment through step-by-step logic tree navigation.
    \item Reducing context window size by focusing on relevant graph information.
    \item Introducing vector-based scoring of semantic similarity, reducing redundant LLM calls.
\end{itemize}

\section{Future Work}
While Conversational AI serves as a compelling case study, the PAF framework holds promise for broader applications. Future research will explore:
\begin{itemize}
    \item \textbf{Node Weights and Path Rules:} Introducing weights and flexible rules.
    \item \textbf{Integration with Different Models:} Experimenting with in-house or domain-specific LLMs.
    \item \textbf{Open-Source Model Improvements:} Tuning embeddings or scoring for domain-specific tasks.
\end{itemize}

\section*{Impact Statement}
This paper presents work whose goal is to advance the field of Machine Learning and Agentic Workflows. There are many potential societal consequences of our work, none which we feel must be specifically highlighted here.

\vspace{1em}
\bibliographystyle{unsrt}

\end{document}